\pdfoutput=1

\documentclass[11pt]{article}

\usepackage{acl}
\usepackage{amsmath,amstext}
\usepackage{amsfonts,amssymb}
\usepackage{times}
\usepackage{latexsym}
\usepackage{graphicx}
\usepackage{booktabs}
\usepackage{multirow}
\usepackage{multicol}
\usepackage{amssymb}
\makeatletter
\renewcommand{\maketag@@@}[1]{\hbox{\m@th\normalsize\normalfont#1}}%
\makeatother

\usepackage{times}
\usepackage{latexsym}
\usepackage[T1]{fontenc}

\usepackage[utf8]{inputenc}

\usepackage{microtype}

\usepackage{inconsolata}

\title{\texorpdfstring{\includegraphics[width=18pt]{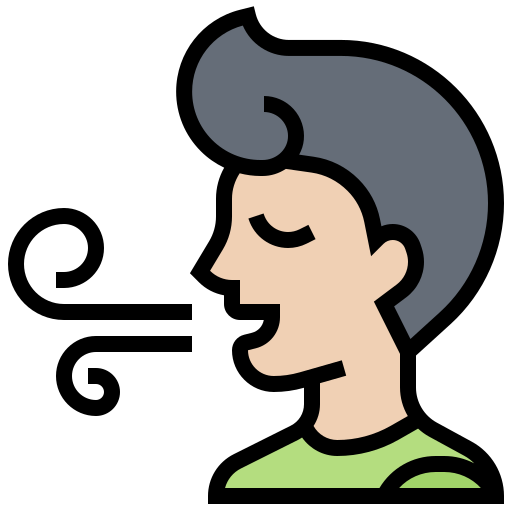}}{}~\textit{Taking a Deep Breath:} Enhancing Language Modeling of\\Large Language Models with Sentinel Tokens}

\author{
\textbf{{Weiyao Luo}\textsuperscript{\rm 1,2,3}, {Suncong Zheng}\textsuperscript{\rm 3}, {Heming Xia}\textsuperscript{\rm 4}, {Weikang Wang}\textsuperscript{\rm 3},} \\
\textbf{{Yan Lei}\textsuperscript{\rm 3,5}, {Tianyu Liu}\textsuperscript{\rm 6}, {Shuang Chen}\textsuperscript{\rm 3} and {Zhifang Sui}\textsuperscript{\rm 1*} } \\
\textsuperscript{\rm 1} State Key Laboratory of Multimedia Information Processing, School of Computer Science, Peking University \\
\textsuperscript{\rm 2} School of Software \& Microelectronics, Peking University ~~
\textsuperscript{\rm 3}Tencent Hunyuan   \\
\textsuperscript{\rm 4}Department of Computing, The Hong Kong Polytechnic University  \\
\textsuperscript{\rm 5}Institute of Computing Technology, Chinese Academy of Sciences ~~
\textsuperscript{\rm 6}Alibaba Group  \\
 {\tt wyluo@stu.pku.edu.cn} ~~ {\tt szf@pku.edu.cn}
}

\begin{document}
\maketitle
\begin{abstract}
Large language models (LLMs) have shown promising efficacy across various tasks, becoming powerful tools in numerous aspects of human life. However, Transformer-based LLMs suffer a performance degradation when modeling long-term contexts due to they discard some information to reduce computational overhead. In this work, we propose a simple yet effective method to enable LLMs to \textit{take a deep breath}, encouraging them to summarize information contained within discrete text chunks. Specifically, we segment the text into multiple chunks and insert special token \texttt{<SR>} at the end of each chunk. We then modify the attention mask to integrate the chunk's information into the corresponding \texttt{<SR>} token. This facilitates LLMs to interpret information not only from historical individual tokens but also from the \texttt{<SR>} token, aggregating the chunk's semantic information. Experiments on language modeling and out-of-domain downstream tasks validate the superiority of our approach.
\end{abstract}

\section{Introduction}
In recent years, Transformer-based large language models (LLMs) have become a focal point of research, leading to the emergence of numerous powerful models such as ChatGPT~\cite{openai2022chatgpt}, GPT-4~\cite{openai2023gpt4}, LLaMA~\cite{llama,touvron2023llama} and Mistral~\cite{jiang2023mistral}. However, in mainstream decoder-only models, subsequent tokens can only attend to preceding historical individual tokens without acquiring information from aggregated local contexts, thereby limiting the language modeling capability of LLMs.


To tackle this problem, existing studies have explored various approaches to compress contexts, such as sentinel tokens~\cite{ren2023context}, memory slots~\cite{ge2023context}, and summary vectors~\cite{zhang2024soaring, chevalier2023adapting}. However, these methods may lose useful context information during the compression process, leading to performance degradation. Besides, some of these compression and accumulation strategies still exhibit quadratic computing complexity of self-attention~\cite{zhang2024soaring}.


\begin{figure}[!t]
\centering
\includegraphics[width=0.96\columnwidth]{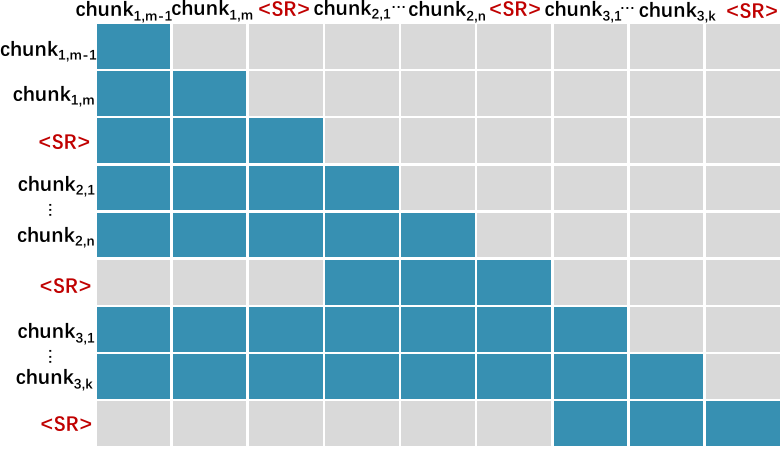} 
\caption{The modified attention mask is illustrated in the figure, where cell $(r, c)$ signifies whether token \textit{r} can attend to token \textit{c}. \(\texttt{chunk}_{2,1}\) represents the first token of the second chunk, with similar patterns for other chunks.}
\label{fig:mask}
\end{figure}

In this work, we introduce a simple yet effective method that allows LLMs to \textit{take a deep breath}, enabling them to gather information not only from preceding historical individual tokens but also from special tokens that encapsulate the holistic information of chunks. Specifically, we propose a strategy to insert new tokens denoted as \texttt{<SR>} (\textbf{S}entinel \textbf{R}ight) at the end of each chunk. During training, the \texttt{<SR>} corresponding to each chunk is capable of attending to the entire content of the chunk. That is to say, after processing each chunk, the LLMs are prompted to summarize key information within this chunk. Consequently, subsequent tokens can acquire information not only from the preceding individual tokens as in the original approach, but also from the sentinel token \texttt{<SR>} which aggregates the holistic information of the chunk. As a result, when generating the next token, tokens can harness both local information (individual tokens within a chunk) and relatively holistic information (the \texttt{<SR>} that represents the entire chunk's context). By prompting the LLMs to take a deep breath, we enable them to obtain richer semantic information during decoding, thereby enhancing its language modeling capability.

We conducted experiments on the Wikitext-2 language modeling benchmark using models ranging from 1.3B to 13B in size, employing diverse positional encoding strategies~\citep{devlin2018bert, su2024roformer}. The experimental results demonstrate that the introduction of sentinel tokens enhances the language modeling capabilities of LLMs. Additionally, we further demonstrated the effectiveness of our method on out-of-domain downstream tasks.

\section{Related Work}

\paragraph{Attention Mask}
A number of architectural modifications have been proposed to constrain and sparsify the attention window~\citep{dai2019transformer, child2019generating}.~\citet{ainslie2023colt5} introduced conditional computation.~\citet{beltagy2020longformer} and~\citet{zaheer2020big} introduced some sparse attention mechanisms to reduce computational complexity. However, most of these architectures require expensive training from scratch. Our approach modifies the attention mask while requiring only a small amount of fine-tuning.

\paragraph{Context Distillation}
\label{sec:context_dis}
Various strategies~\cite{askell2021general, snell2022learning} have been proposed for prompt compression and context distillation.~\citet{ren2023context} adopted a random partitioning approach to compress random contiguous tokens into a single token, resulting in a significant performance degradation. The AutoCompressors~\citep{chevalier2023adapting} compress context into summary vectors, exhibiting quadratic complexity.~\citet{mu2023learning} compressed instructions into short prefixes, a method similar to the memory slots introduced by~\citet{ge2023context}, which may lead to the loss of some useful information. Our method prompts LLMs to take a deep breath, yielding richer semantic information.

\section{Approach}
In this section, we will present our approach which integrating the comprehensive information of each chunk into the sentinel \texttt{<SR>}. This is achieved by strategically placing sentinel token \texttt{<SR>} at the right flank of a chunk and modifying the attention mask rules accordingly.

\subsection{Adding Sentinel Tokens}
Assuming there is a text segment that has been divided into multiple chunks, to enable subsequent tokens to extract information not only from individual tokens in the preceding text but also from the collective information from the aggregated semantic content of a chunk, we introduce the special sentinel token: \texttt{<SR>}, which represents sentinel right. The sentinel token is inserted at the end of a chunk to mark its boundary, and it is also added to the model's vocabulary.

Specifically, to absorb the information of a chunk into the \texttt{<SR>} sentinel, we implement a modified causal attention mask, as illustrated in Figure~\ref{fig:mask}. The strategy for ordinary tokens (excluding \texttt{<SR>}) is the same as that of a standard causal attention mask, where they can attend to all preceding tokens. For \texttt{<SR>}, which is the crux of our method, to enable it to encapsulate the semantic information of the corresponding chunk, \texttt{<SR>} can attend to the ordinary tokens within the chunk. Through this modification of the mask strategy, in conjunction with fine-tuning, we promote the condensation of semantic information from a chunk of tokens into the sentinel token at the end, allowing \texttt{<SR>} to become an aggregator capable of selecting and retrieving information from the corresponding chunk.

\begin{table*}[!ht]
  \centering
  \small
  \caption{Perplexity (the lower, the better) of six LLMs on the WikiText-2 language modeling benchmark.}
  \label{results}
  \begin{tabular}{l|cccccc}
    \toprule
        \multirow{1}{*}{Method} & \multicolumn{1}{c}{OPT-1.3B} & \multicolumn{1}{c}{OPT-2.7B} & \multicolumn{1}{c}{RedPajama-3B} & \multicolumn{1}{c}{Mistral-7B} & \multicolumn{1}{c}{Llama-2-7b} & \multicolumn{1}{c}{Llama-2-13b} \\ 
        \midrule
    Origin   & 14.044   & 12.416   & 10.557       & 6.408      & 6.235      & 5.709  \\
    Sentinel & \textbf{12.664}   & \textbf{11.313}   & \textbf{9.387}        & \textbf{6.176}      & \textbf{6.01}       & \textbf{5.512}  \\
    
    \bottomrule
  \end{tabular}
\end{table*}

\subsection{Adapting Model Inputs for Sentinel Integration}
\label{sec:sentinel_integration}

For the general next token prediction task, the computation of the loss relies on labels corresponding to each token. With the introduction of sentinel tokens, these labels must be adjusted accordingly. Specifically, if the current token is a sentinel token, identified by \texttt{<SR>}, the label for the subsequent position should be uniquely designated to ensure that the location corresponding to the current sentinel token is excluded from the loss calculation.

Conversely, if the next token is a sentinel, its label should be set to that of the subsequent non-sentinel token. Moreover, the position ids of sentinel tokens should be congruent with the position id of the last non-sentinel token preceding the current position.


\section{Experiments}
\label{sec:typestyle}
\subsection{Models and Data}
\label{ssec:subhead}
To validate the effectiveness of our method, we selected models of various sizes and position encoding methods, including OPT-1.3B, OPT-2.7B~\cite{zhang2022opt}, RedPajama-3B~\cite{together2023redpajama}, Mistral-7B~\cite{jiang2023mistral}, Llama-2-7B, and Llama-2-13B~\cite{touvron2023llama}.

We choose the Wikitext-2 dataset~\cite{merity2016pointer}, which is composed of Wikipedia articles and widely used for evaluating language modeling. By fine-tuning on Wikitext-2, we report perplexity (PPL) on the test set as an evaluation metric. It is noteworthy that sentinel tokens are not included in the computation of PPL, as discussed in Section~\ref{sec:sentinel_integration}.

\subsection{Experimental Setup}


We treat each individual sentence as a chunk, more details in \ref{sec:breathAnalysis}. For training, the LoRA~\citep{hu2021lora} technique was adopted for fine-tuning during training, with all parameters of the LLM frozen except for the embeddings of the special sentinel tokens and the LoRA matrices. In our approach, the LoRA module is applied to all attention layers, typically comprising the q\_proj, k\_proj, v\_proj, and o\_proj parts. The rank of the LoRA is set to 16. We used the AdamW~\citep{loshchilov2018fixing} with a learning rate of 5e-5. The batch size is set to 12. The entire experiment, including modifications to the attention mask, was based on the Huggingface transformers library~\citep{wolf2020transformers}.

\begin{figure}[!t]
\centering
\includegraphics[scale=0.3]{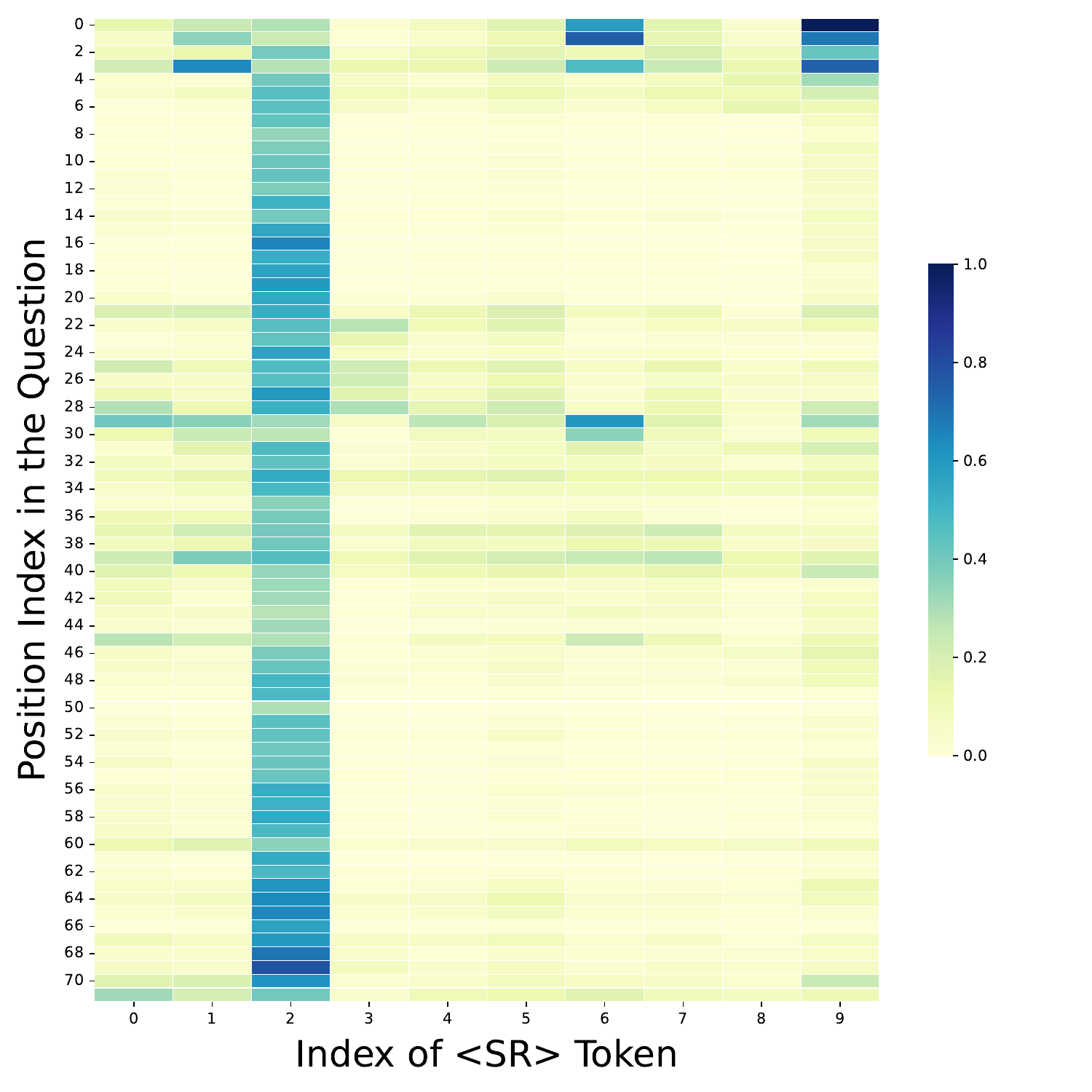} 
\caption{An example of DocumentQA illustrating the attention between each position in the question and sentinel tokens, where correct \texttt{<SR>} index is 2. More detailed explanation provided in Section~\ref{sec:out_analysis}.}
\label{fig:attention}
\end{figure}



\begin{table*}[!ht]
  \centering
  \small
  \caption{Perplexity (the lower, the better) when a chunk contains different numbers of sentences.}
  \label{tab:breath_results}
  \begin{tabular}{c|cccccc}
    \toprule
        \multirow{1}{*}{\#Sentences} & \multicolumn{1}{c}{OPT-1.3B} & \multicolumn{1}{c}{OPT-2.7B} & \multicolumn{1}{c}{RedPajama-3B} & \multicolumn{1}{c}{Mistral-7B} & \multicolumn{1}{c}{Llama-2-7b} & \multicolumn{1}{c}{Llama-2-13b} \\ 
        \midrule
    1 & \textbf{12.664}   & \textbf{11.313}   & \textbf{9.387}        & \textbf{6.176}      & \textbf{6.01}       & \textbf{5.512}  \\
    2   & 13.637   & 12.113   & 10.026     & 6.438     & 6.278      & 5.756  \\
    3   & 14.078   & 12.509   & 10.309     & 6.550     & 6.412      & 5.852  \\
    4   & 14.181   & 12.605   & 10.385     & 6.591      & 6.404      & 5.863  \\
    \bottomrule
  \end{tabular}
\end{table*}

\subsection{Results}
The experimental results are shown in Table~\ref{results}, where \textbf{Origin} represents the results of the original model after standard fine-tuning without any special modifications, and \textbf{Sentinel} denotes the results fine-tuning the model after adding the special sentinel token \texttt{<SR>}. Notably, the methods in Section~\ref{sec:context_dis} center on context compression, trading off information for reduced computation and, consequently, lower performance. Given their predictable inferiority to \textbf{Origin}, it is superfluous to include these results in Table~\ref{results}. Our work stands as a pioneering exploration of this method in language modeling.


It can be observed that our method achieved notable performance enhancements across all models, as evidenced by a reduction in perplexity. In the OPT series models and the RedPajama-3B model, perplexity significantly decreased by approximately 10\% compared to the vanilla method. In the Llama series and Mistral-7B, perplexity also decreased by about 3.5\%.

\subsection{Analysis}
\label{sec:analysis}
These results suggest that our approach effectively assists LLMs with the ability in acquiring information from diverse perspectives during next token prediction. By prompting LLMs to take a deep breath, they not only obtain information from each individual token, similar to general LLMs, but also capture more holistic information. This is achieved through the use of the special sentinel token \texttt{<SR>}, which represents the entire chunk of information. By incorporating, the model can take a deep breath to access richer semantic information during decoding, thereby enhancing its language modeling capability.

\paragraph{Breath Length Analysis}
\label{sec:breathAnalysis}
To explore the optimal interval for taking a breath to achieve the best results, we conducted experiments with chunks containing 1 to 4 sentences, as shown in Table~\ref{tab:breath_results}. The findings indicate that the best performance is achieved when the chunk contains only one sentence, and a decrease in performance is observed across all models as the number of sentences increases. Therefore, in the main results presented in Table~\ref{results}, we adopted the strategy of allowing the model to take a deep breath after every sentence to achieve the best effect.




\subsection{Generalization of Model Performance on Out-of-Domain}

\subsubsection{Data}
\label{sec:datadetails}
We select 1,105 samples of the MLQA dataset~\cite{lewis2019mlqa} for DocumenQA task evaluation. Additionally, we choose 1,120 samples of MultiNews~\cite{alex2019multinews} to assess the model's performance on summarization. More details can be found in Appendix~\ref{app:datadetails}.

\subsubsection{Models and Metrics}
We utilize OPT-1.3B and OPT-2.7B \cite{zhang2022opt}, which were previously fine-tuned on the Wikitext-2 dataset, for out-of-domain DocumentQA and summarization tasks, respectively. The evaluation follows LongBench \cite{bai2023longbench} methodologies. For DocumentQA, the model generates an output based on the given context and question, and the F1 score is calculated by comparing the predictions with ground-truth references. For summarization, the model generates summaries from a simple prompt and the provided article, with Rouge-L scores \cite{lin2004rouge} computed between the generated and reference summaries.



\begin{table}[t]
    \centering
    \small
    \begin{tabular}{l|c c|c c}
    \toprule
    \multicolumn{1}{c|}{\multirow{2}{*}{Task}} & \multicolumn{2}{c|}{OPT-1.3B} & \multicolumn{2}{c}{OPT-2.7B} \\
    \cmidrule(lr){2-3} \cmidrule(lr){4-5}
    & Origin & Sentinel & Origin & Sentinel \\
    \midrule
    DocumentQA & 3.97 & \textbf{19.89} & 17.41 & \textbf{21.78} \\
    Summarization & 2.77 & \textbf{7.69} & 7.58 & \textbf{11.02} \\
    \bottomrule
    \end{tabular}
    \caption{Experimental results for OPT-1.3B and OPT-2.7B on two out-of-domain tasks. We show the F1-Score for the DocumentQA task and Rouge-L for the Summarization task, respectively.}
    \label{tab:DocumentQA_and_summary_results}
\end{table}




\subsubsection{Out-of-domain Results and Analysis}
\label{sec:out_analysis}
Table~\ref{tab:DocumentQA_and_summary_results} presents the experimental results, showing that compared to the original models, adding sentinels significantly improved the performance on both out-of-domain tasks, particularly for OPT-1.3B on the DocumentQA task, where it nearly quadrupled the F1 score. Additionally, the larger OPT-2.7B with more parameters outperformed the smaller OPT-1.3B, which is consistent with the scaling law~\cite{kaplan2020scaling}. These findings further validate the effectiveness and robustness of our approach across various out-of-domain tasks.

Figure~\ref{fig:attention} illustrates the attention distribution of the question sequence towards the special token \texttt{<SR>} in a DocumentQA task case. In this example, the document contains 10 chunks, each followed by an \texttt{<SR>} token. Therefore, the horizontal axis displays the indices of these \texttt{<SR>} tokens, ranging from 0 to 9. In this DocumentQA example, the question segment has a length of 72, as indicated on the vertical axis. This means that the vertical axis represents the index of each position in the question, and each position corresponds to the attention scores for the 10 \texttt{<SR>} tokens in the document. 

The index corresponding to the chunk that contains the true answer to the question is 2. The figure reveals that nearly every position within the question segment exhibits a heightened attention value for the token \texttt{<SR>} at index 2. This suggests that the token \texttt{<SR>} encapsulates the semantic information of its corresponding chunk, and this information can be accurately captured during decoding, thereby improving the model's performance.


\section{Conclusions}
\label{sec:conclusion}
In this work, we introduce a novel approach that prompts LLMs to take a deep breath after encountering each chunk. This strategy enables LLMs not only to extract information from individual tokens but also to be capable of selecting and retrieving information from aggregators(denoted as \texttt{<SR>}) corresponding to each chunk. Experiments on language modeling and out-of-domain downstream tasks demonstrate the effectiveness of our method.



\section*{Limitations}
\label{sec:limitations}
In this work, we only prompt Large Language Models (LLMs) to take a deep breath on OPT models with up to 2.7 billion parameters, RedPajama model with 3 billion parameters, the Mistral-7B, and Llama2 models with up to 13 billion parameters. Future endeavors should be directed towards establishing the efficacy of this approach on models of an even greater magnitude. Furthermore, the exploration of a broader range of potential chunk division strategies presents a valuable avenue for further research.


\section*{Acknowledgements}
Weiyao Luo and Zhifang Sui are supported by the National Key Research and Development Program of China 2020AAA0106700.

\bibliography{anthology,custom}

\clearpage
\appendix

\section*{Appendix}

\section{More Data Details}
\label{app:datadetails}
For the document question answering (DocumentQA) task, we selected the MLQA~\citep{lewis2019mlqa} dataset for evaluation. MLQA is a multilingual QA task dataset, from which we choose the English portion as the test set for our task. 

DocumentQA refers to a task where a model is provided with a \textbf{Document}, which contains several sentences, along with a corresponding \textbf{Question}. The expectation is that the model can process the \textbf{Document} and output the correct answer to the question. Evaluation is conducted by comparing the model-generated results with the ground-truth labels to judge the quality of the model's output.

For the summarization task, we chose the MultiNews~\cite{alex2019multinews} dataset and filtered it based on example length, ultimately selecting 1120 data points for the evaluation of the summarization task. The original MultiNews dataset includes over 5k examples, but we find that the length of some examples might exceed the context length that our evaluation model can handle, so we ultimately selected 1120 examples for testing.





\end{document}